\documentclass[11pt]{article}

\usepackage[final]{acl}

\usepackage{times}
\usepackage{latexsym}
\usepackage{graphicx}

\usepackage[T1]{fontenc}

\usepackage[utf8]{inputenc}

\usepackage{microtype}

\usepackage{inconsolata}

\usepackage{seqsplit}
\usepackage{float}
\usepackage{etoolbox}
\usepackage{hyperref}

\title{AAbAAC: An Annotated Corpus for Autoimmunity Information Extraction}

\author{
Fabien Maury\textsuperscript{1,2}, 
Solène Grosdidier\textsuperscript{3}, 
Maud de Dieuleveult\textsuperscript{1,*}, 
Adrien Coulet\textsuperscript{2,*}\\[1em]
   \textsuperscript{1}Inserm, Universit\'e Paris Cit\'e, U1163 Institut Imagine, Paris, France\\
   \textsuperscript{2}Inria, Inserm, Universit\'e Paris Cit\'e, U1346 HeKA, Paris, France\\
   \textsuperscript{3}Freelance researcher, The Hague, Netherlands\\[1em]
   {
    $^{*}$These authors contributed equally,  
   Correspondence: fabien.maury@inserm.fr
   }
}

\begin{document}
\maketitle

\begin{abstract}

Despite advances in information extraction driven by deep learning and large language models,  performance gaps remain in highly specialized biomedical fields, where domain-specific complexity poses challenges for generalist models.
In this work, we focus on the domain of autoimmunity, where the main entities of interest are autoimmune diseases, autoantibodies (\textit{i.e.}, molecules that may mark or cause these diseases), their molecular targets, their location in the body, and their associated clinical signs. 
Herein, we present AAbAAC (AutoAntibodies and Autoimmunity Annotated Corpus), a corpus of 115 abstracts selected from PubMed, where we manually annotated entities and their relationships. 
First, AAbAAC was used to evaluate several methods on the task of named entity recognition (NER), and secondly, to fine-tune NER models. 
Our study demonstrates the utility of AAbAAC for information extraction in the domain of autoimmunity, showing expected improvement in NER performance after fine-tuning. This illustrates the value of small-scale annotation efforts for specialized domains and contributes to the computational study of autoimmunity. The AAbAAC corpus is available at 
\href{https://github.com/f-maury/AAbAAC}{https://github.com/f-maury/AAbAAC}.
\end{abstract}

\section{Introduction}
Autoimmune diseases arise from dysfunction of the adaptive immune system, whereby molecules of the self are targeted by self-antibodies, named autoantibodies. Those are generally detected in blood or cerebrospinal fluid, and widely used as a biomarker of autoimmune diseases, critically guiding differential diagnosis. Autoimmune diseases collectively affect an estimated 3–5\% of the global population~\citep{ahsan_origins_2023, shapira_defining_2010}. Yet, when considered individually, many of these diseases are rare~\cite{hayter_updated_2012}. 
In part due to the rarity of these pathologies, there is no single centralized resource including all relevant information about autoimmunity. Indeed, relevant bits of domain expertise are scattered across various sources, including many recent research papers. For this reason, information extraction is a relevant way to study autoimmune diseases, by automating the processing of large numbers of documents into structured, interoperable, and machine-readable knowledge bases. 

Lately, BERT-based models have consistently achieved state-of-the-art performance in information extraction tasks, particularly when adapted through fine-tuning on domain-specific documents matching the target domain.
While many domain-specific models exist, high-quality annotated data to adapt general models to highly specialized or niche domains are also lacking in many cases.  
To our knowledge, this is the case in human autoimmunity, where no manually annotated corpus dedicated to autoimmune diseases and autoantibodies exists to evaluate or fine-tune models specialized for the recognition of autoantibodies, autoimmune diseases, and their relationships. To fill this gap, we propose AAbAAC, a corpus of PubMed abstracts annotated for entities and relations related to autoimmunity, and demonstrate its value for named entity recognition (NER).

Section 2 reviews related works. Section 3 describes the methodology used to create the corpus, and Section 4 presents the resulting corpus. Section 5 introduces the experimental setup for named entity recognition (NER) using AAbAAC, and Section 6 reports the associated results. Finally, the paper concludes with perspectives for future work enabled by this new corpus.

\section{Related works}

\subsection{Existing resources for autoimmunity}

Some general and widely used knowledge bases and resources exist for broad biomedical use, such as the Human Phenotype Ontology (HPO)~\cite{gargano_human_2024}, Orphanet~\cite{rath_representation_2012}, or the UMLS~\cite{bodenreider_unified_2004}.
Those include information about autoimmune diseases and autoantibodies, but are not sufficiently exhaustive, detailed, nor up-to-date for real-world studies. Indeed, autoimmunity is a specialized and evolving topic that is not the focus of generalist resources. They can nonetheless be used to generate an initial list of autoantibody names, that can then be used for information extraction.

More specialized resources have also been developed that include information about antibodies, autoantibodies, or related elements. For example, this is the case of IEDB~\cite{vita_immune_2025}, a database about epitopes, \textit{i.e.}, regions of antigens bound by antibodies, or AAgAtlas~\cite{wang_aagatlas_2017}, which focuses on autoantigens. Each of these resources focuses on a topic that is relevant to the study of autoimmunity, but none of them compile all the relevant knowledge in the field of autoimmunity for clinicians or researchers to use. However, for information extraction purposes, these resources could be used to build lists of autoantibody names from their antigen names.

Some annotated text corpora already exist for biomedical information extraction, but to the best of our knowledge, none is dedicated to human autoimmunity and autoimmune diseases. For example, the MedMentions corpus~\cite{mohan_medmentions_2019} tackles the biomedical domain in a general manner,  with annotations of UMLS terms of most Semantic Types. As a result, it may include the most common autoimmune diseases and autoantibodies, but it does not include uncommon ones that do not correspond to UMLS terms, nor does it include relationships between the entities. The ABAG corpus~\cite{dinh_extract_2022} focuses on antibodies and antigens, but is not specific to autoimmunity and does not provide annotations of relationships between entities. Therefore, existing resources are of limited use for autoimmunity-related information extraction, since isolating in these resources the annotations specific to autoimmunity would be challenging and result in limited coverage of autoimmunity-related concepts.

\subsection{NER for autoimmunity}
Many works make use of dictionaries and rule-based methods to conduct NER with decent performance. Such dictionaries can be built from either ontologies or databases. For example, a recent work by \citet{remaki_improving_2025} used such a combination, including terms from SNOMED CT~\cite{wang_snomed_2002} and rules, to extract biology exams and drugs in relations to immune-mediated inflammatory diseases.

Another work by \citet{subramanian_antibody_2017} extracts antibody names using simple regular expressions based on the fact that antibodies are often written as "X antibody", or "antibody to X", where "X" is the target of the antibody. This method does not require a dictionary of possible target names, but fails at covering the various possible ways an antibody can be mentioned in text. Moreover, in the context of autoimmunity, such a method would struggle to differentiate autoantibodies from regular antibodies.

The development of multi-head attention and transformer-based architectures of language models enables efficient information extraction from long texts with more flexibility and context awareness than rule-based approaches. For NER tasks, encoder models such as BERT-based models~\cite{devlin_bert_2019} perform competitively while being lighter and less computationally expensive to work with than large language models (LLM)~\cite{naguib-etal-2024-shot}. The original BERT model has been adapted to many domains, including the biomedical domain. However, we did not find any model specialized for the recognition of autoimmunity-related entity types.

\section{Corpus creation methods}
\subsection{Text selection}
\label{sec:text_selection}


To select a set of relevant texts to be annotated, we first built a dictionary of autoantibody names and synonyms from HPO terms, which we used to query a PubMed API~\cite{sayers_e-utilities_2018} for titles and abstracts of scientific papers. Indeed, HPO includes terms describing positivity to some autoantibody detection tests (terms under \textit{HP:0030057}). Most of them are of the form: "\textit{anti-X antibody positivity}", where "\textit{anti-X antibody}" is the name of an autoantibody that targets the antigen "\textit{X}". We relied on this regular syntax to extract names of autoantibodies and manually reviewed results. 
We also generated lexical variants by reducing each autoantibody name to a core string through the removal of selected prefixes and suffixes, then recombining this core with alternative affixes. For example, "\textit{anti-smooth muscle antibody}" was reduced to "\textit{smooth muscle}" by removing "\textit{anti}" and "\textit{antibody}", and one resulting variant was "\textit{smooth muscle autoantibody}". Overall, the dictionary includes a total of 10,916 variations for 285 unique autoantibodies that originated from HPO release 2024-07-01. On average, each autoantibody in the dictionary has 38.30 variations, including synonyms that were already listed in the HPO.



For each autoantibody type in the dictionary, PubMed is searched with a query of the form "\textit{(""autoantibody\_name\_1""[Title/Abstract] OR ""autoantibody\_name\_2""[Title/Abstract] AND humans[MeSH Terms] AND (antibodies[MeSH Terms] OR autoantibodies[MeSH Terms])}". MeSH term filters were added to select only papers tagged with "human" and either "antibodies" or autoantibodies. This process returned a total of 56,750 titles and abstracts. 
The full query we used is available in the project's GitHub repository.


Despite MeSH terms filtering, some of the obtained texts were irrelevant due to not being written in English or not being related to autoimmunity. For this reason, we enforced additional filters: abstracts were automatically pre-annotated with exact matches of autoantibody names from our HPO-derived dictionary, and with disease and autoantibody names identified by GLiNER. This pre-annotation was performed with a version of the GLiNER specialized for the biomedical field ("\textit{urchade/gliner\_large\_bio-v0.1}").
This model is an early version of GLiNER large that is finetuned on PubMed abstracts. 


Only texts containing at least one HPO autoantibody term match, or one disease pre-annotation and one autoantibody pre-annotation were kept, reducing the set of texts to 44,890.

Among these, 120 texts were randomly selected for manual annotation, out of which 5 were manually eliminated due to either not being related to autoantibodies or not being English texts, resulting in a final set of 115 texts. The initial selection size was 120 because we aimed to have a corpus of around 100 texts and expected a few of them to be eliminated.

\subsection{Annotation guidelines and process}

Annotation was performed by 4 annotators with a background in either medical informatics, biology, or pharmacy. This group elaborated annotation guidelines through two preliminary annotation batches to identify main entity types, difficulties, and edge cases before the start of the annotation campaign~\cite{hovy_towards_nodate}. 
Annotation guidelines are available in the following GitHub repository: \href{https://anonymous.4open.science/r/aabaac-FA3F}{https://anonymous.4open.science/r/aabaac-FA3F}.
Five types of entities were annotated: "\textit{Autoantibody}", "\textit{Autoantibody target}", "\textit{Disease}", "\textit{Symptom or clinical sign}", and "\textit{Autoantibody location}"; and ten different types of relationships: "\textit{is associated with}", "\textit{is not associated with}", "\textit{may be associated with}", "\textit{is caused by}", "\textit{is target of}", "\textit{is reference to}", "\textit{discontiguous entity}" (this relation is used to link several separated parts of a single entity), "\textit{is synonym of}", "\textit{is located in}", and "\textit{is subclass of}".

To make the annotation task easier for annotators, it was agreed to not annotate obvious relationships between nested entities, which were automatically added after manual annotation. For instance, when a "\textit{Target}" is found inside the span of an "\textit{Autoantibody}", which typically happens in structures like "\textit{anti-X antibody}", the annotators were instructed not to add an "\textit{is target of}" relation. Similarly, when an "\textit{Autoantibody}" is found inside the span of a related "\textit{Symptom or clinical sign}", the annotators were instructed not to add an "\textit{is associated with}" relationship. Those were automatically added after manual annotation.

\begin{figure}[t]
    \centering
    \includegraphics[width=\linewidth]{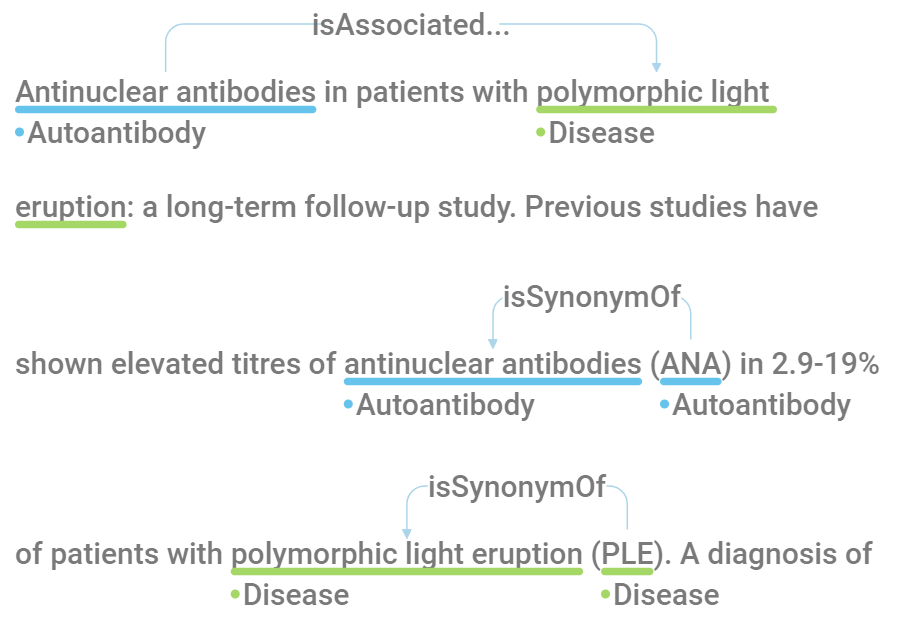}
    \caption{Example of text annotated in the Doccano web interface.}
    \label{fig:doccano-interface}
\end{figure}

Each text was annotated independently by 2 annotators, but no single annotator annotated all texts. Each annotator was paired with each other annotator with equal frequency, and each pair annotated the same number of texts. 
Annotation was performed using Doccano~\cite{doccano} on texts where entities had been pre-annotated, but not relations. These pre-annotations were obtained by string matching and GLiNER. Doccano web interface is illustrated in \autoref{fig:doccano-interface}. 

Texts were sent to the annotators in small numbered batches of around 4 texts, and annotators were free to annotate them in any order, in a single or multiple sessions, and at any time before the end of the campaign.
Finally, an adjudication took place where annotations from both annotators for each text were confronted, and a consensual final annotation was decided by the head annotator.

\section{Resulting corpus description}

This section includes descriptive counts and statistics from the final adjudicated corpus of 115 texts.
The variability of the length of the texts is high (from 50 to 3600 characters, as per \autoref{aggregate-stats}), as is their annotation content (from 2 to 88 entities, and from 0 to 55 relations). This is partially due to the fact that some of the texts are limited to an article title with an empty abstract. 

\begin{table}[H]
  \centering
  \small
  \setlength{\tabcolsep}{4pt}
  \begin{tabular}{lrrrr}
    \hline
    \textbf{Metric} & \textbf{Min} & \textbf{Max} & \textbf{Average} & \textbf{Median} \\
    \hline
    Entities per text   & 2  & 88   & 29.18   & 27 \\
    Relations per text  & 0  & 55   & 15.47   & 14 \\
    Text length (chars) & 50 & 3600 & 1500.94 & 1500 \\
    Sentences per text  & 1  & 64   & 13.46   & 12 \\
    \hline
  \end{tabular}
  \caption{\label{aggregate-stats}
    General corpus statistics.
  }
\end{table}

Imbalance is also observed in the number of entities of each type, with especially only 64 (see \autoref{entity-stats}) occurrences of "\textit{Autoantibody location}", whereas \textit{Disease} is the most common entity with 1159 occurrences. 

\begin{table}[H]
  \centering
  \small
  \setlength{\tabcolsep}{4pt}
   \begin{tabular}{lrrr}
    \hline
    \textbf{Entity type} & \textbf{Count} & \textbf{Avg/text} & \textbf{Median/text} \\
    \hline
    Autoantibody               & 891  & 7.75  & 6 \\
    Autoantibody location      & 64   & 0.56  & 0 \\
    Autoantibody target        & 581  & 5.05  & 3 \\
    Disease                    & 1159 & 10.08 & 8 \\
    Symptom or clinical sign   & 661  & 5.75  & 5 \\
    \hline
    \textbf{Total} & 3356 & 29.18 & 27 \\
    \hline
  \end{tabular}
  \caption{\label{entity-stats}
    Counts and distributions in the corpus by entity type.
  }
\end{table}

Extremely variable numbers of occurrences can be observed for each relation type. Most relation annotations consist of "\textit{is associated with}" (735, see \autoref{relation-stats}), which can be explained by its rather broad definition, "\textit{is target of}" (494), and "\textit{is synonym of}" (226). Some relation types have almost never been used, such as "\textit{is caused by}" (13), which could be explained by the cautious tone used in scientific literature, and the difficulty to establish causation with certainty, and "\textit{is reference to}" (2), which could be explained by its constrained definition in our guidelines and redundancy with "\textit{is synonym of}".

\begin{table}[H]
  \centering
  \small
  \setlength{\tabcolsep}{4pt}
  \begin{tabular}{lrrr}
    \hline
    \textbf{Relation type} & \textbf{Count} & \textbf{Avg/text} & \textbf{Median/text} \\
    \hline
    discontiguousEntity & 106 & 0.92 & 0 \\
    isAssociatedWith    & 735 & 6.39 & 5 \\
    isCausedBy          & 13  & 0.11 & 0 \\
    isLocatedIn         & 73  & 0.64 & 0 \\
    isNotAssociatedWith & 24  & 0.21 & 0 \\
    isRefTo             & 2   & 0.02 & 0 \\
    isSubClassOf        & 61  & 0.53 & 0 \\
    isSynonymOf         & 226 & 1.97 & 1 \\
    isTargetOf          & 494 & 4.30 & 3 \\
    mayBeAssociatedWith & 45  & 0.39 & 0 \\
    \hline
    \textbf{Total} & 1779 & 15.47 & 14 \\
    \hline
  \end{tabular}
  \caption{\label{relation-stats}
    Counts and distributions in the corpus by relationship type.
  }
\end{table}

To evaluate the difficulty of the annotation task, inter-annotator agreement measures were computed between each pair of annotators, and between each annotator and the final annotations. Since the annotators simultaneously had to decide whether a text contained annotations, where in the text were these annotations located, if any, and of what type they were, Mathet's gamma coefficient~\citep{mathet-etal-2015-unified, Titeux2021} was used. Mathet's gamma is a measure that takes into account both disagreement on span limits (unitizing) and on label type (categorizing) for entity annotations.
To measure agreement on relation annotations, three different comparison functions (strict match, flexible with tolerance for some disagreement, and super-flexible with tolerance for more disagreement) were used to compute F1-scores between the annotations of each annotator pair. We report a general average gamma coefficient between pairs of annotators of 0.67. The average gamma was 0.74 for "\textit{Autoantibody}", 0.71 for "\textit{Autoantibody location}", 0.44 for "\textit{Autoantibody target}", 0.81 for "\textit{Disease}", and 0.54 for "\textit{Symptom or clinical sign}". 

Relationship annotations are intrinsically more complex, and this is particulary the case in this work where annotators had 10 different relation types to choose from. To assess this difficulty quantitatively, we measured the inter-annotator agreement for relationships using an F1-score. We obtain a F1-score of 0.37 for all relation types together, using a custom flexible comparison metric tolerating some small overlap differences, and some entity type confusions.

Considering all relation types jointly, the primary source of disagreement was \textit{isAssociatedWith}, because it was the most frequently used relation type and because its broad definition allowed for subjective interpretation. When examining relation types individually, some exhibited lower agreement scores than \textit{isAssociatedWith}; however, their contribution to the overall disagreement remained limited due to their lower frequency. In particular, \textit{isRefTo} was used only twice and obtained an F1 score of 0. Given its limited use, removing this relation type from the annotation scheme could be considered. Overall, the relatively low inter-annotator agreement observed for relations highlights the difficulty of annotating texts from a specialized scientific domain. Furthermore, the complexity of the annotation scheme made the annotation task challenging for annotators.


\section{Evaluating NER methods}

\subsection{Experimental setup}

To demonstrate the utility of the AAbAAC corpus in autoimmunity NER, several methods were compared.



The corpus was randomly divided into train and test following an 80 - 20 split. This procedure was repeated independently five times, resulting in five distinct train–test partitions. Supervised approaches were fine-tuned separately on each training set, and all approaches were evaluated on the corresponding test sets. Reported results are the average performance across the five test sets.


Chunking of longer texts occurred after the split so that all chunks from the same text remain part of the same set. This prevents information leakage from the train set to the test set.

\subsection{QuickUMLS}

The first method tested is QuickUMLS~\cite{soldaini_quickumls_2016} on the task of recognition of mentions of "\textit{Autoantibody}" and "\textit{Disease}" entities only. One dictionary was created for each of these entity types. The dictionary for "\textit{Disease}" was created by selecting all the UMLS terms from Semantic Type "\textit{Disease or Syndrome}", and only keeping English strings associated with these terms. The "\textit{Autoantibody}" dictionary was by combining the dictionary made from an intial list of HPO terms for text selection (see Subsection \ref{sec:text_selection}) and was then enriched from LOINC \cite{forrey_logical_1996} and manual additions. 
Moreover, name variations for autoantibodies were automatically added to the dictionary. QuickUMLS was used with default parameters except for the threshold, which was set to 0.9.

\subsection{GLiNER}

Although standard BERT-based models achieve state-of-the-art performance in NER, they rely on predefined annotation schemas and task-specific fine-tuning, in contrast to the GLiNER architecture, which supports more flexible, label-agnostic entity recognition~\cite{zaratiana_gliner_2024}. GLiNER does not need a fixed list of entity types, but can work with any label passed as input at inference time. While studying a BERT model would be of interest in our use case, we decided to work with the GLiNER architecture (GLiNER 2.1 models: \href{https://huggingface.co/urchade/gliner_large-v2}{urchade/gliner\_large-v2}) on the task of NER for entity types related to autoimmunity, notably autoantibodies. Indeed, this allows for more flexibility with regards to the set of entity types to be identified, and our annotation scheme was initially not fixed. In this 
setup, texts longer than 384 tokens were chunked by cutting at the end of the last sentence before reaching the maximum length, and each chunk was tokenized using GLiNER's tokenizer.

GLiNER base models (small, medium, and large) were evaluated in a zero-, two-shot, and fine-tuned setting. 
In the GLiNER architecture, the labels are passed to the model along with the input text at inference time. The way labels are formulated may affect performances, and it may be useful to reformulate them or use synonyms to see what returns best results (e.g., "\textit{Autoantibody target}" or "\textit{Autoantigen}"). At this step, labels that seemed the most concise and clear were picked, but except from light changes, we did not conduct an extensive experiment to search for potential optimal alternatives. 
GLiNER models were evaluated on the task of NER for the following entities of interest: "\textit{Autoantibody}", "\textit{Autoantibody location}", "\textit{Autoantibody target}", "\textit{Disease}", "\textit{Symptom or clinical sign}", on the 5 different test sets obtained from our split of the AAbAAC corpus.  These entity types are the strings that were passed to GLiNER along with texts, as part of the input.

For the 2-shot setting, we passed 2 short, manually made-up examples of each entity type before the actual input text. The examples followed the format: '"sentence": "The patient was admitted after complaining of thigh pain."\string\n"Symptom or clinical sign": "thigh pain"'. The same 2 example sentences for each type were passed along each input text. 
In rare cases, the input text was long and close to the maximal allowed chunk length, and adding the extra examples has led to truncation, which may have affected the performance slightly. For the 2-shot experiment, the GLiNER threshold parameter, which acts as a confidence filter for the model’s output was set to 0.3, but for 0-shot and fine-tuned configuration it was set to 0.5. 

GLiNER base models were fine-tuned using the train set from each split, and the fine-tuned models were evaluated on the associated test set. 
For fine-tuning, the tokenized texts of the train set are shown to the model along with their entity annotations. For evaluation, the chunked but untokenized texts of the test set are passed to the model along with the types of entities to identify.

\subsection{MedGemma}

Google's MedGemma model for text (\href{https://huggingface.co/google/medgemma-27b-text-it}{google/medgemma-27b-text-it}) was evaluated in 0-shot, 2-shot setting and finally fine-tuned using AAbAAC.
MedGemma being a generative model, the NER task was framed as a task where the model had to output a structured JSON answer containing identified entities, as a response to a prompt containing the input text and instructions. The LLM was instructed to write a JSON string containing the text spans identified as entities. The LLM's output was then parsed to extract valid JSON segments and exclude malformed output parts. The generated answer was limited to a length of 1024 tokens. The model was not asked to return the exact character offsets delimiting the spans, as preliminary tests with this method seemed unreliable. For this experiment, the chunks of texts were given at train time, untokenized, as input to the model along with corresponding entities. For evaluation, the chunks of texts were also passed untokenized to the model along with instructions, in a prompt.
The fine-tuning was conducted using rank-16 LoRA adapters, with the 4-bit quantized version of the model.

\section{Results}

This section presents the averaged results for the various NER methods we evaluated. 

Results of the QuickUMLS experiments can be found 
in table \autoref{quickumls-results}. It can be observed that performances are higher for the "\textit{Disease}" label than for the "\textit{Autoantibody} label. This could be explained by the fact that for the recognition of "\textit{Disease}" we relied on the richness of UMLS vocabularies: our disease dictionary is 636,989 rows, whereas our manual dictionary of 
"\textit{Autoantibody}" is smaller (7,464 rows) and handcrafted, and thus probably incomplete, which would explain the lower recall for autoantibodies. 

\begin{table}[H]
  \centering
  \small
  \setlength{\tabcolsep}{4pt}
  \begin{tabular}{lccc}
    \hline
    \textbf{Entity type} & \textbf{Precision} & \textbf{Recall} & \textbf{F1 ($\pm$SD)}     \\
    \hline
    ALL            & 0.47 & 0.40 & 0.43 ($\pm 0.04$)\\
    Autoantibody   & 0.50 & 0.24 & 0.33 ($\pm 0.06$)\\
    Disease        & 0.46 & 0.53 & 0.49 ($\pm 0.06$)\\
    \hline
  \end{tabular}
  \caption{\label{quickumls-results}
    Precision (P), recall (R), and F1-score (F1) of QuickUMLS for autoantibodies and diseases, using custom made dictionaries.
  }
\end{table}

Table \ref{model-comparison} reports the performance of all the different model and experimental setup we evaluated. 
Overall best F1-score is obtained with the fine-tuned version of MedGemma. We particularly observe that in terms of F1-score, QuickUMLS performance is not significantly lower than GLiNER and MedGemma if those are not fine-tuned. F1-score for entity type "\textit{ALL}" is the average over 5 test sets of the micro F1-score computed by counting true positives, false positives, and false negatives of all entity types together.

Regarding the various GLiNER configurations, the fine-tuned largest version of GLiNER is overall performing slightly better than other versions. 2-shot GLiNER models perform overall worse than 0-shot, possibly indicating our entity labels were somewhat more confusing than helpful to the model without a sufficient number of examples. Top precision is achieved by GLiNER large 2-shot, whereas top recall and F1-score are achieved by GLiNER large fine-tuned (\autoref{model-comparison}). All methods return higher precision than recall, and this seems to be especially the case in the 2-shot configurations. It should be possible to modulate this to some extent by acting on the GLiNER threshold parameter, as well as on the number and type of examples. 

\begin{table}[H]
  \centering
  \small
  \setlength{\tabcolsep}{4pt}
  \begin{tabular}{llccc}
    \hline
    \multicolumn{2}{c}{\textbf{Model}} & \textbf{P} & \textbf{R} & \textbf{F1 ($\pm$SD)} \\
    \hline
    \hline
    \multicolumn{2}{c}{QuickUMLS}           & 0.47 & 0.40 & 0.43 ($\pm 0.04$)\\
    \hline
    &GLiNER small       & 0.72 & 0.26 & 0.39 ($\pm 0.05$)\\
0-shot&GLiNER medium    & 0.67 & 0.33 & 0.45 ($\pm 0.04$)\\
    &GLiNER large       & 0.74 & 0.30 & 0.42 ($\pm 0.03$)\\
    &MedGemma           & 0.61 & 0.31 & 0.41 ($\pm 0.05$)\\
    \hline
    &GLiNER small       & 0.76 & 0.20 & 0.31 ($\pm 0.05$)\\
2-shot&GLiNER medium    & 0.67 & 0.26 & 0.37 ($\pm 0.04$)\\
    &GLiNER large       & \textbf{0.80}   & 0.21 & 0.34 ($\pm 0.04$)\\
    &MedGemma           & 0.67   & 0.29 & 0.41 ($\pm 0.04$)\\
    \hline
    &GLiNER small       & 0.65 & 0.51  & 0.58 ($\pm 0.05$)\\
fine-&GLiNER medium     & 0.67 & 0.52 & 0.58 ($\pm 0.03$)\\
tuned&GLiNER large      & 0.67 & 0.53 & 0.59 ($\pm 0.04$)\\
    &MedGemma           & 0.71 & \textbf{0.62} & \textbf{0.66} ($\pm 0.03$)\\
    \hline
  \end{tabular}
  \caption{\label{model-comparison}
    Precision (P), recall (R), and F1-score (F1) of different models for all entity types together.
  }
\end{table}

We also report in Table \ref{label-performance-2} per-label performances for some of the best-performing models: GLiNER large and MedGemma, both in 0-shot and fine-tuned configurations. Fine-tuning increases model F1-score for all entity types, though it sometimes slightly decreases precision: performance gains are mostly due to recall gains. Considerable performance imbalance can be observed between labels: for the fine-tuned models, the highest F1-score (0.79) is achieved by GLiNER large for the detection of "\textit{Disease}", as per \autoref{label-performance-2}; and the lowest F1-score is achieved by MedGemma for the detection of "\textit{Autoantibody location}" (0.33). In the case of "\textit{Autoantibody location}", fine-tuning GLiNER large causes 
performance increase from an F1-score of 0.07 to 0.60; and fine-tuning MedGemma increases F1-score from 0.01 to 0.33. It is possible that some types of entities, such as "\textit{Autoantibody location}" are not easily captured when the only information about them is their label, without examples. However, giving examples can lead to better performances, as illustrated by the performance differences between 0-shot and fine-tuned configurations.

\begin{table}[t]
  \centering
  \small
  \setlength{\tabcolsep}{4.4pt}
  \begin{tabular}{p{3.6cm}ccc}
    \hline
    \textbf{Model} and Entity type & \textbf{P} & \textbf{R} & \textbf{F1 ($\pm$SD)} \\
    \hline
    \hline
    \textbf{GLiNER large 0-shot} & & & \\
    \ \ ALL                         & 0.74 & 0.30 & 0.42 ($\pm 0.03$)\\
    \ \ Autoantibody                & 0.72 & 0.37 & 0.48 ($\pm 0.05$)\\
    \ \ Autoantibody location       & 0.10 & 0.06 & 0.07 ($\pm 0.05$)\\
    \ \ Autoantibody target         & 0.76 & 0.06 & 0.11 ($\pm 0.07$)\\
    \ \ Disease                     & 0.82 & 0.48 & 0.60 ($\pm 0.06$)\\
    \ \ Symptom or clinical sign    & 0.56 & 0.13 & 0.21 ($\pm 0.06$)\\
    \hline
    \textbf{GLiNER large fine-tuned} & & & \\

    \ \ ALL                      & 0.67 & 0.53 & 0.59 ($\pm 0.04$)\\
    \ \ Autoantibody             & 0.75 & 0.49 & 0.59 ($\pm 0.11$)\\
    \ \ Autoantibody location    & 0.47 & 0.88 & \textbf{0.60} ($\pm 0.17$)\\
    \ \ Autoantibody target      & 0.43 & 0.28 & 0.34 ($\pm 0.07$)\\
    \ \ Disease                  & 0.80 & 0.77 & \textbf{0.79} ($\pm 0.02$)\\
    \ \ Symptom or clinical sign & 0.52 & 0.39 & \textbf{0.42} ($\pm 0.05$)\\
        \hline

    \textbf{MedGemma 0-shot} & & & \\
    \ \ ALL                      & 0.61 & 0.31 & 0.41 ($\pm 0.05$)\\
    \ \ Autoantibody             & 0.65 & 0.45 & 0.53 ($\pm 0.06$)\\
    \ \ Autoantibody location    & 0.02 & 0.01 & 0.01 ($\pm 0.03$)\\
    \ \ Autoantibody target      & 0.42 & 0.07 & 0.11 ($\pm 0.02$)\\
    \ \ Disease                  & 0.70 & 0.39 & 0.50 ($\pm 0.06$)\\
    \ \ Symptom or clinical sign & 0.47 & 0.23 & 0.31 ($\pm 0.05$)\\
        \hline  
    \textbf{MedGemma fine-tuned} & & & \\

    \ \ ALL                      & 0.71 & 0.62 & \textbf{0.66} ($\pm 0.03$)\\
    \ \ Autoantibody             & 0.79 & 0.70 & \textbf{0.74} ($\pm 0.06$)\\
    \ \ Autoantibody location    & 0.35 & 0.35 & 0.33 ($\pm 0.18$)\\
    \ \ Autoantibody target      & 0.75 & 0.67 & \textbf{0.70} ($\pm 0.04$)\\
    \ \ Disease                  & 0.79 & 0.66 & 0.71 ($\pm 0.02$)\\
    \ \ Symptom or clinical sign & 0.46 & 0.41 & \textbf{0.42} ($\pm 0.04$)\\
    \hline
  \end{tabular}
  \caption{\label{label-performance-2}
    Precision (P), recall (R), and F1-score (F1) per label for GLiNER large and MedGemma fine-tuned, both in 0-shot and fine-tuned configuration. Best F1-score per type of entity is reported in bold.
  }
\end{table}

\section{Discussion}

This work demonstrates the use of the AAbAAC corpus for autoimmunity information extraction through some basic NER experiments, including model fine-tuning. Indeed, models fine-tuned using the AAbAAc corpus perform better at detecting entities related to autoimmunity than the same models used out-of-the-box, or than a rule-based approach. However, so far, in the work presented here, no search for optimal fine-tuning parameters was conducted. The fine-tuned models' performances could probably be improved by performing a grid search or some other type of optimization prior to fine-tuning.

The presented AAbAAC corpus aims to be of use for complete information extraction pipelines for the study of autoimmunity. This includes both named entity recognition and extraction of relations between these entities, so the corpus also includes relation annotations. However, for now, no evaluation of the use of the corpus for information extraction was conducted. 
One future work may use the GLiNER2~\cite{zaratiana_gliner2_2025} architecture, which enables the joint extraction of both entities and relations simultaneously.

For GLiNER-based models, so far, only fine-tuning of the original generalist GLiNER family models was performed, however it might be of interest to attempt similar experiments with different GLiNER variants. For instance, versions dedicated to the biomedical field, such as \textit{Ihor/gliner-biomed-large-v1.0} \cite{yazdani2025gliner} are available.
Attempting to fine-tune a domain-specific BERT model, such as PubMedBERT~\cite{gu_domain-specific_2022}, for example, would also be a possible future work direction and an interesting comparison.

\section{Conclusion}

This work introduces AAbAAC, a manually annotated corpus of texts that was created 
for information extraction in the field of autoimmunity. 
AAbAAC includes annotations of entities and relationships that are relevant to the study of autoantibodies and autoimmune diseases. Preliminary NER experiments on this new corpus demonstrate that fine-tuning models with AAbAAC yields a performance increase over dictionary-based approaches as well as zero-shot or few-shot settings.
Future work will focus on further tuning different models for this domain, for both tasks of NER and relation extraction.

\section{Limitations}

AAbAAC corpus is relatively small: 115 annotated texts of variable length, all of them being titles and abstracts of scientific papers published on PubMed. This is enough to deliver substantial performance gains on NER tasks for the entity types considered in this work; however, ultimately the knowledge that can be derived from this corpus is finite and may not reflect the entire existing spectrum of autoantibodies, autoimmune diseases, and the various appellations used in the literature to refer to these concepts, especially since many autoimmune diseases are rare~\cite{miller_increasing_2023}. For some very rare, or unusually named autoantibodies not annotated in the AAbAAC corpus, detection improvements may be limited. In addition, since this corpus consists of texts drawn from the scientific literature, its impact on NER in clinical texts could be lower due to differences between written styles and possibly different naming conventions for autoantibodies.

Regarding relation annotations, some of the relation types in our scheme have almost never been instantiated in the corpus. For example, this is the case with \textit{isCausedBy} (13 occurrences, see \autoref{relation-stats}), a stronger relation than the more common \textit{isAssociatedWith} (735 occurrences), or \textit{isRefTo} (2 occurrences).

The annotation process took place using the Doccano tool, which does not allow to specify annotation attributes. Therefore, to annotate entities that are composed of several discontiguous segments of text, the \textit{discontiguousEntity} relation is used. But this complexity is lost to an NER model that only considers entity annotations, and this may have caused some performance drop in the presented experiments.

Other models from the state of the art could have been interesting to experiment with as well, in particular BERT-type models fine-tuned for the biomedical literature, or non-proprietary LLM. Their inclusion could open perspectives to extend our study to consider the computational cost of various approaches and attempt to specialize or distill efficient but costly models. 

\section*{Acknowledgments}

The authors acknowledge the Fili\`eres de Sant\'e Maladies Rares BRAIN-TEAM and Filnemus for funding. The authors thank B. Belloir and S. Bernichtein from BRAIN-TEAM and R. Soussi from Filnemus for their help. The authors highlight the contributions of the BRAIN-TEAM, Filnemus, FAI²R, Fimarad, Fimatho, FILFOIE, MHEMO, and DéfiScience networks. The authors also thank C. Lucano and C. Fabrizzi (ORPHANET) for their help and our research teams, HeKA and Pathophysiological basis of skeletal dysplasia, for their constant support.

\bibliographystyle{acl_natbib}
\bibliography{custom}

\end{document}